\documentclass[10pt,twocolumn,letterpaper]{article}

\usepackage{cvpr}              
\usepackage{array}
\usepackage[accsupp]{axessibility}

\usepackage[dvipsnames]{xcolor}

\definecolor{cvprblue}{rgb}{0.21,0.49,0.74}
\usepackage[pagebackref,breaklinks,colorlinks,citecolor=cvprblue]{hyperref}
\usepackage{color,multirow}

\def\paperID{14877} 
\def\confName{CVPR}
\def\confYear{2024}

\title{Learning Spatial Adaptation and Temporal Coherence in Diffusion Models \\
	   for Video Super-Resolution\thanks{{\small This work was performed at HiDream.ai.}}}

\author{\normalsize Zhikai Chen$^{\dag}$, Fuchen Long$^{\S}$, Zhaofan Qiu$^{\S}$, Ting Yao$^{\S}$, Wengang Zhou$^{\dag}$, Jiebo Luo$^{\ddag}$, and Tao Mei$^{\S}$\\
	$^{\dag}$\normalsize MoE Key Laboratory of Brain-inspired Intelligent Perception and Cognition, University of Science and Technology of China \\ 
	$^{\ddag}$\normalsize University of Rochester, Rochester, NY USA \quad
	$^{\S}$\normalsize HiDream.ai Inc. \\
	{\tt\small\ czk654@mail.ustc.edu.cn}, {\tt\small\{longfuchen, qiuzhaofan, tiyao\}@hidream.ai} \\
	{\tt\small\ zhwg@ustc.edu.cn}, {\tt\small\ jluo@cs.rochester.edu}, {\tt\small\ tmei@hidream.ai} \\
}

\begin{document}

\maketitle

\begin{abstract}
Diffusion models are just at a tipping point for image super-resolution task. Nevertheless, it is not trivial to capitalize on diffusion models for video super-resolution which necessitates not only the preservation of visual appearance from low-resolution to high-resolution videos, but also the temporal consistency across video frames. In this paper, we propose a novel approach, pursuing Spatial Adaptation and Temporal Coherence (SATeCo), for video super-resolution. SATeCo pivots on learning spatial-temporal guidance from low-resolution videos to calibrate both latent-space high-resolution video denoising and pixel-space video reconstruction. Technically, SATeCo freezes all the parameters of the pre-trained UNet and VAE, and only optimizes two deliberately-designed spatial feature adaptation (SFA) and temporal feature alignment (TFA) modules, in the decoder of UNet and VAE. SFA modulates frame features via adaptively estimating affine parameters for each pixel, guaranteeing pixel-wise guidance for high-resolution frame synthesis. TFA delves into feature interaction within a 3D local window (tubelet) through self-attention, and executes cross-attention between tubelet and its low-resolution counterpart to guide temporal feature alignment. Extensive experiments conducted on the REDS4 and Vid4 datasets demonstrate the effectiveness of our approach.
\end{abstract}

\begin{figure}[t]
	\centering
	\includegraphics[width=0.99\linewidth]{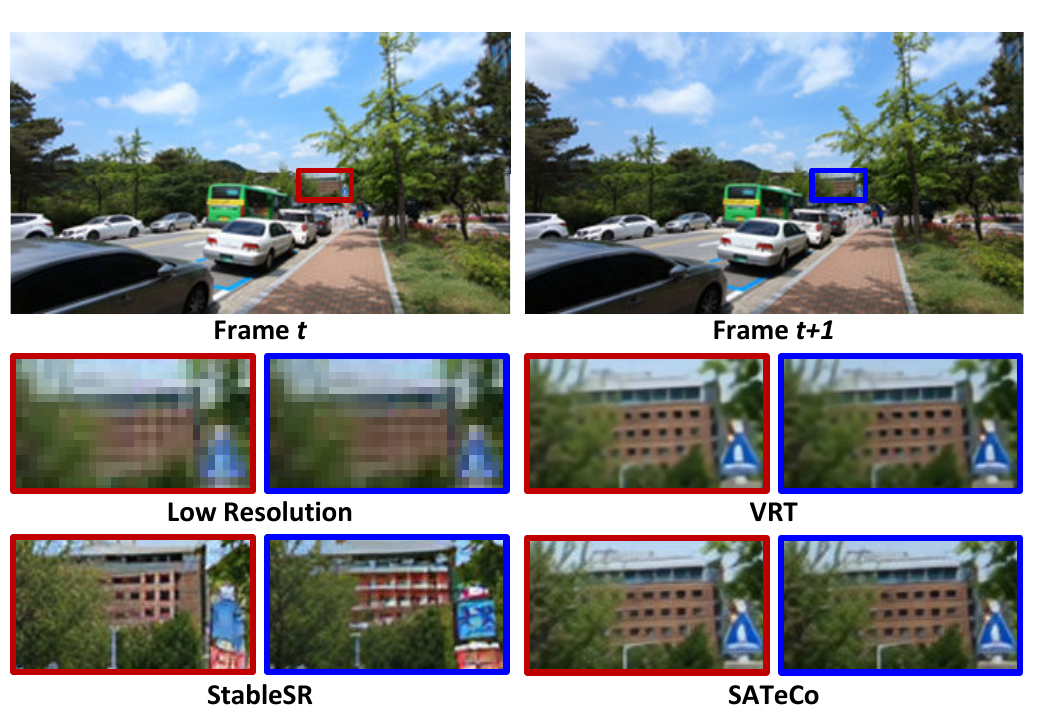}
	\vspace{-0.10in}
	\caption{An illustration of video super-resolution by using different approaches of StableSR \cite{wang2023stablesr}, VRT \cite{liang2022vrt} and our SATeCo to generate two adjacent frames. The region in the same local position is presented in the zoom-in view.}
	\label{fig:intro}
	\vspace{-0.2in}
\end{figure}

\section{Introduction}

In recent years, diffusion models \cite{rombach2022high,zhang2023adding,dhariwal2021diffusion,saharia2022palette} have shown great progress in revolutionizing image generation. In between, a series of image super-resolution works \cite{wang2023stablesr,yang2023pasd,rombach2022high} benefit from leveraging knowledge prior embedded in diffusion models to upscale low-resolution (LR) images into high-resolution (HR) ones. Compared to 2D images, videos have one more time dimension, bringing more challenges when capitalizing on diffusion models for video super-resolution (VSR). One natural way is to utilize the pre-trained diffusion models for image super-resolution (ISR), e.g., StableSR \cite{wang2023stablesr}, to magnify each video frame. The representative advances \cite{wang2023stablesr, yang2023pasd} manifest that diffusion models for ISR could synthesize more details than traditional regression models, e.g., VRT \cite{liang2022vrt}. As depicted in Figure \ref{fig:intro}, the edges of the windows in the building produced by StableSR are much clearer than those generated by VRT. Nevertheless, the inherent stochasticity of diffusion models might jeopardize the spatial fidelity and hallucinate some extra visual content. Moreover, the independent frame-wise super-resolution overlooks the relation across consecutive frames, resulting in the issue of frame inconsistency in the high-resolution videos. For instance, the traffic signs in Figure \ref{fig:intro} are totally different between the two adjacent frames generated by StableSR.

In general, the difficulty of exploring diffusion models for video super-resolution originates from two aspects: 1) how to alleviate the stochasticity in diffusion process to preserve visual appearance? 2) how to guarantee the temporal consistency across frames in the HR videos?   
We propose to address the two issues through learning spatial-temporal guidance from low-resolution videos to manage diffusion procedure for video super-resolution.
To regulate spatial adaptation, we estimate affine parameters on the LR frame features to modulate each pixel in HR frames. As such, the pixel-wise guidance is employed to nicely learn the feature of every pixel in HR frames and better improve spatial fidelity. In an effort to temporally cohere video frames, we strengthen feature interaction across HR frames, and feature calibration between HR frames and LR counterpart via the attention mechanism. Moreover, large receptive field is attained by conducting the self-attention and cross-attention on the features within a 3D local window (tubelet), thereby facilitating temporal feature alignment.

To materialize our idea, we present a new SATeCo method to carry out Spatial Adaptation and Temporal Coherence for video super-resolution. 
Technically, SATeCo uses a transformer-based video upscaler to up-sample the input LR video.  
The VAE encoder then extracts the video features and latent code of LR video, which are further exploited for diffusion calibration. 
SATeCo deliberately devises spatial feature adaptation (SFA) and temporal feature alignment (TFA) modules, and inserts the two modules into each decoder block of UNet and VAE, for latent-space video denoising and pixel-space video reconstruction. 
In the regularization of latent-space video denoising, SFA exploits two convolutional layers on the latent code of each up-sampled LR frame, to predict a scale and bias to modulate the pixel-wise feature of HR frame. 
TFA first executes self-attention on HR video latent code within a tubelet to enhance feature interaction, and further performs cross-attention between the tubelet and its LR counterpart for feature calibration in HR video. 
The LR video features are exploited in the same way to regulate the HR video feature learning in pixel-space video reconstruction.
SATeCo finally refines the decoded HR video by referring to the up-sampled LR video via a neural network to balance synthesized quality and fidelity.

The main contribution of this paper is the proposal of SATeCo to explore spatial adaptation and temporal coherence in diffusion models for video super-resolution.
The solution also leads to the elegant views of how to leverage pixel-wise information from LR videos for visual appearance preservation, and how to achieve frame consistency in HR video generation.
Extensive experiments on REDS4 and Vid4 verify the superiority of SATeCo in terms of both spatial quality and temporal consistency.

\begin{figure*}[t]
	\centering
	\includegraphics[width=0.95\linewidth]{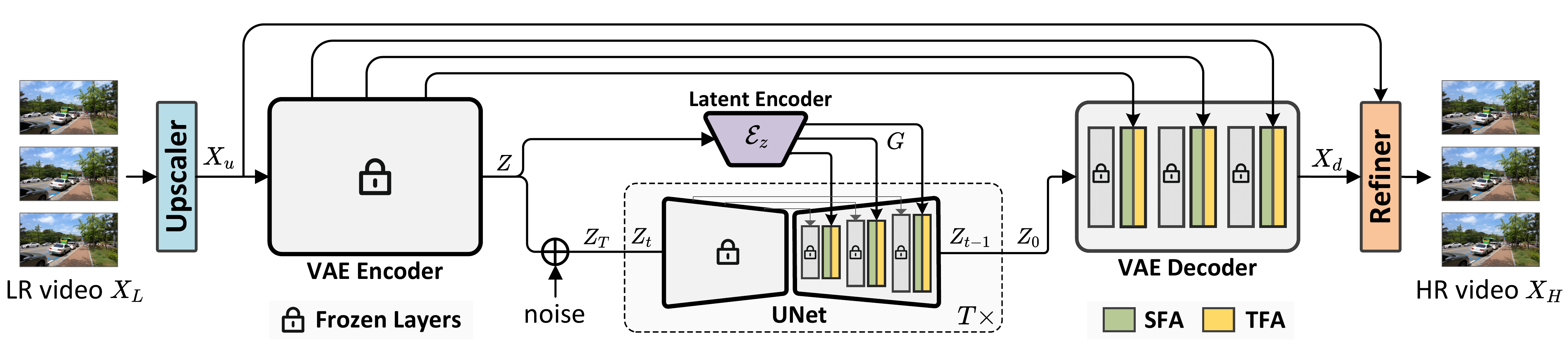}
	\caption{
		An overview of our SATeCo architecture. The input LR video $X_L$ is first up-sampled to the target resolution via a transformer-based video upscaler. Then, the up-sampled video $X_u$ is fed into the VAE encoder to extract the video features and latent code $Z$. 
		Next, the Gaussian noise is added into $Z$ according to the diffusion scheduler, and the noisy video latent code is then restored by UNet for quality enhancement.
		In latent space, a latent encoder extracts the LR latent feature maps $G$ on the LR latent code $Z$, followed by spatial feature adaptation (SFA) and temporal feature alignment (TFA) modules in each decoder block of UNet for spatial-temporal guidance learning. 
		Given the denoised video latent code $Z_0$, the VAE decoder decodes the video $X_d$ based on the guidance learnt by SFA and TFA on LR video features. 
		Finally, the decoded video $X_d$ is adjusted by a video refiner via referring to $X_u$ for final HR video $X_H$ synthesis. 
	}
	\label{fig:framework}
	\vspace{-0.2in}
\end{figure*}

\section{Related Work}

\textbf{Video super-resolution.} 
Modern VSR approaches are mainly based on deep neural networks and can be grouped into two categories, i.e., sliding window-based methods and recurrent methods.
Early sliding window-based VSR techniques \cite{xue2019flow,caballero2017real,li2020mucan,xu2021temporal,yi2019progressive} rely on 2D or 3D CNNs \cite{isobe2020tga,jo2018deep} which incorporate a sequence of LR frames to predict center HR frame.
To fully utilize the complementary information across adjacent frames, the deformable convolutions \cite{edvr2019,tian2020tdan} are employed for feature alignment.
Inspired by the success of transformer architecture in various computer vision tasks \cite{long2022dynamic,long2022sifa,chen2023anchor,long2023pc}, self-attention emerges to be integrated into the VSR frameworks \cite{wang2020deep,liu2022learning,liang2022vrt,geng2022rstt}.
One representative is VRT \cite{liang2022vrt} which plugs the temporal mutual attention block into transformer backbone to facilitate motion estimation, feature alignment and fusion.  
Nevertheless, the sliding window-based approaches are difficult to capture long-range dependencies which could limit the performance of video super-resolution.
In contrast to aggregate information from adjacent frames in a short term, recurrent approaches~\cite{chan2021basicvsr,chan2022basicvsr++,shi2022rethinking,liang2022recurrent,huang2017video,isobe2020video,haris2019recurrent,sajjadi2018frame,yi2021omniscient} utilize a hidden state to sequentially propagate information from all previous frames to the current frame, benefiting the frame restoration.
For instance, Chan \emph{et al.} \cite{chan2021basicvsr} adopt a bidirectional propagation scheme with flow-based feature alignment to maximize information gathering in super resolution.
Despite having the great capacity of the recurrent models for temporal information gathering, the local details are still hard to be restored when the LR video encounters significant degradation in a long temporal range.

\textbf{Diffusion models for super-resolution.} 
Impressive performances of image synthesis achieved by diffusion models \cite{dhariwal2021diffusion,zhang2023adding, choi2021ilvr,alexander2022glide,hertz2022prompt,videodrafter} encourage the deployment on image super-resolution.
These explorations \cite{kawar2022denoising, wang2022zero, zhu2023restoration, fei2023generative, chung2022improving,chung2022diffusion,meng2022diffusion,song2022pseudoinverse} leverage the knowledge prior embedded in the pre-trained diffusion models to magnify images.
For example, StableSR \cite{wang2023stablesr} integrates a time-aware encoder into Stable-Diffusion \cite{rombach2022high} model without altering the pre-trained weights, and achieves promising super-resolution results.
To further enhance the reconstruction of image texture details, Yang \emph{et al.} \cite{yang2023pasd} introduce an attention-based control module to maintain pixel consistency between LR and HR images.
Different from the advances which optimize a small part of inserted parameters, several approaches \cite{kawar2022denoising,fei2023generative,wang2022zero} fix all weights in the pre-trained synthesis model and attempt to incorporate constraints into the reverse diffusion process to guide image restoration.
Although the effectiveness of knowledge prior has been manifested in various diffusion-based ISR methods, it is still a grand challenge to employ diffusion models for video super-resolution and preserve spatial fidelity and temporal consistency.

In summary, our work mainly focuses on diffusion models for video super-resolution.
The proposal of SATeCo contributes by exploring not only how to preserve spatial fidelity through modulating HR frame features, but also how to calibrate HR video features with LR counterpart for better temporal feature alignment.

\section{Our Approach}

In this section, we present our newly-minted SATeCo, pursuing Spatial Adaptation and Temporal Coherence in diffusion models for video super-resolution. 
Figure \ref{fig:framework} depicts an overview of the architecture. SATeCo begins with a video upscaler to increase the resolution of the input LR video. Then, the up-sampled video is fed into VAE encoder for video feature extraction and latent code prediction. 
After that, a spatial feature adaptation (SFA) and a temporal feature alignment (TFA) module are leveraged to learn spatial-temporal guidance on latent code and features of LR video, to calibrate latent-space video denoising and pixel-space video reconstruction. 
As such, the two modules are plugged into each block of the decoder in UNet and VAE. 
In procedure of video latent code denoising, SFA estimates the affine parameters on LR video latent code to modulate each pixel of the HR video latent code. 
TFA first performs self-attention on the HR video latent code within a tubelet, and further enhances latent code by executing cross-attention between the tubelet and its LR counterpart. 
Similarly, SFA and TFA are conducted in the VAE decoder to guide HR video reconstruction with the LR video features. 
Finally, SATeCo designs a video refiner to adjust the decoded HR video by referring to the up-sampled video for a good trade-off between synthesized quality and fidelity. 

\subsection{Video Upscaler}

Most existing VSR approaches \cite{xue2019flow,shi2022rethinking} first upscale the input LR videos through a resampling operation, and then improve their visual quality. 
Nevertheless, the widely adopted resampling operations, e.g., Bilinear and Bicubic sampling, might damage the original visual patterns \cite{shi2022rethinking} in LR frames, having a negative impact on the subsequent video enhancement. 
Therefore, we exploit the recipe of reducing frame degradation ahead of the feature learning \cite{chan2022investigating} in neural networks and propose a video upscaler, which generates more accurate up-sampled videos for the following quality enhancement by diffusion models. 

\begin{figure*}[t]
	\centering
	\includegraphics[width=0.95\linewidth]{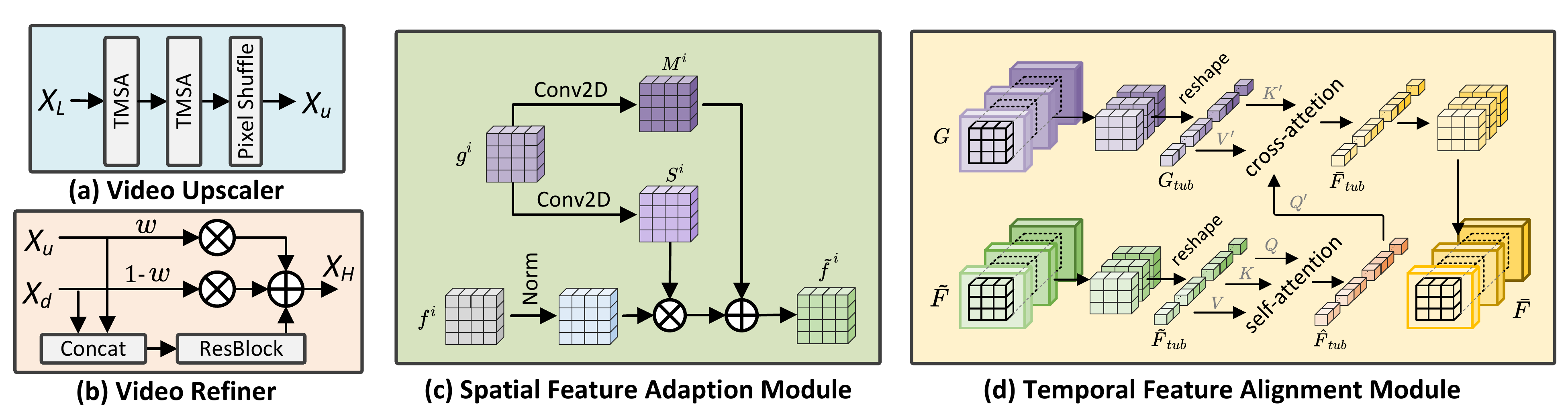}
	\caption{An illustration of (a) video upscaler, (b) video refiner, (c) spatial feature adaptation and (d) temporal feature alignment module.}
	\label{fig:blocks}
	\vspace{-0.1in}
\end{figure*}

Given the input LR video $X_L$, we utilize a transformer-based video upscaler for video up-scaling as illustrated in Figure \ref{fig:blocks}(a). 
It consists of two cascaded temporal mutual self-attention (TMSA) blocks \cite{liang2022vrt} to temporally aggregate video features, and a pixel-shuffle layer \cite{shi2016real} to increase video spatial resolution via feature reshaping. 
The up-sampled video $X_u=\{x_u^i\}_{i=1}^L$ with $L$ frames is then fed into the diffusion model for video quality enhancement. 

\subsection{Spatial Feature Adaptation Module}
The inherent stochasticity \cite{yang2023pasd} of diffusion models might result in the distortion of texture details in image super-resolution.
A natural way of employing diffusion models for super-resolution is to learn the spatial-level condition via convolution-based \cite{wang2023stablesr} or transformer-based \cite{yang2023pasd} structure to guide latent code denoising in UNet.
Such kind of mechanism only manages feature regularization in latent space, posting difficulty to learn sufficient inductive bias and provide precise guidance for high-resolution image restoration. The similar issue also exists in video super-resolution. 
To alleviate this, we introduce a spatial feature adaptation (SFA) module which dynamically learns pixel-wise guidance from the input LR videos for diffusion calibration. In the meanwhile, the SFA module emphasizes the inductive bias learning in both of the latent-space video denoising (i.e., training of UNet) and pixel-space video reconstruction (i.e., training of VAE).

Figure \ref{fig:blocks}(c) illustrates our SFA module. 
Given the up-sampled LR video $X_u$, the VAE encoder first encodes $X_u$ into the video latent code $Z = \{z^i\}_{i=1}^L$.
Next, we exploit a convolution-based latent encoder $\mathcal{E}_z$ to extract the LR latent feature maps $G=\mathcal{E}_z(Z)$, which are further utilized to guide the HR feature learning in UNet decoder.
Formally, we denote the HR intermediate feature maps in UNet and the LR latent feature maps in latent encoder as $F=\{f^i\}_{i=1}^L$ and $G=\{g^i\}_{i=1}^L$, respectively.
For the $i$-th frame, we measure a scale ratio $S^i \in \mathbb{R}^{{H}\times{W}\times{C}}$ and a bias $M^i \in \mathbb{R}^{{H}\times{W}\times{C}}$ for each pixel of HR intermediate feature map $f^i \in \mathbb{R}^{{H}\times{W}\times{C}}$ based on the LR latent feature map $g^i \in \mathbb{R}^{{H}\times{W}\times{C}}$ via two 2D convolution layers:
\begin{equation}
M^i = \mathrm{Conv2D}(g^i),\quad S^i = \mathrm{Conv2D}(g^i).
\end{equation} 
Then, the output HR feature map $\tilde{f}^i$ in UNet is generated by modulating the normalized HR intermediate feature map $f^i$ with $S^i$ and $M^i$ as:
\begin{equation}
\tilde{f}^i = S^i \odot \frac{f^i - \mu^i}{\sigma^i} + M^i,
\end{equation}
where $\odot$ denotes point-wise multiplication. $\mu^i$ and $\sigma^i$ are the mean and standard deviation values of the feature map $f^i$.
Hence, the affine parameters estimated on the latent feature maps of LR videos calibrate the intermediate feature maps of HR videos in latent code denoising, which adaptively injects the pixel-wise information into the video latent code to preserve the visual appearance. 
For video feature learning in pixel space, SFA module is inserted into each block of VAE decoder.
Similarly, the extracted video features of LR videos are taken as the guidance to estimate the affine parameters in SFA module to adjust HR video feature learning for video reconstruction.
We take all the modulated intermediate feature maps $\tilde{f}^i$ from SFA module as $\tilde{F} = \{\tilde{f}^i\}_{i=1}^L$, which is employed for the following temporal feature alignment in the UNet and VAE decoders.

\setlength{\tabcolsep}{1.1mm}{
	\begin{table*}\small
		\caption{Performance comparisons in terms of pixel-based (PSNR and SSIM) and perception-based (LPIPS, DISTS, NIQE and CLIP-IQA) evaluation metrics on the REDS4 and Vid4 datasets. The width and height of the LR videos are rescaled by $4$ times through different VSR approaches. We follow VRT \cite{liang2022vrt} to set the frame number as $6$ in each clip for HR video inference.} 
		\vspace{-0.2cm}
		\centering
		\begin{tabular}{c||l|c|c|c|c|c|c|c||c}
			\toprule[1pt]
			{Datasets} & Metrics & Bicubic & StableSR~\cite{wang2023stablesr} & TOFlow~\cite{xue2019flow} & EDVR-M~\cite{edvr2019}  & BasicVSR~\cite{chan2021basicvsr} & VRT~\cite{liang2022vrt} & IconVSR~\cite{chan2021basicvsr} & SATeCo \\
			\hline			
			\multirow{5}{*}{REDS4} & PSNR$\uparrow$ & 26.14 & 24.79 & 27.98 & 30.53 & 31.42 & 31.60 & \textbf{31.67} & \underline{31.62} \\
			&SSIM$\uparrow$ & 0.7292 & 0.6897 & 0.7990 & 0.8699 & 0.8909 & 0.8888 & \textbf{0.8948} & \underline{0.8932}\\
			&LPIPS$\downarrow$ & 0.3519 & 0.2412 & 0.3104 & 0.2312 & 0.2023 & 0.2077& \underline{0.1939} & \textbf{0.1735} \\
			&DISTS$\downarrow$ & 0.1876 & \underline{0.0755} & 0.1468 &0.0943 & 0.0808 & 0.0823 & {0.0762} & \textbf{0.0607} \\
			&NIQE$\downarrow$ & 7.257 & \underline{4.116} & 6.260 & 4.544 & 4.197 & 4.252 & 4.117 & \textbf{4.104} \\
			&CLIP-IQA$\uparrow$ & 0.6045 & \underline{0.6579} & 0.6176 & 0.6382 & 0.6353 & 0.6379 & 0.6162 & \textbf{0.6622} \\
			\midrule
			\multirow{5}{*}{Vid4} & PSNR$\uparrow$ & 23.78 & 22.18 & 25.89 & 27.10 & 27.24 & \textbf{27.93} & 27.39 & \underline{27.44}\\
			&SSIM$\uparrow$ & 0.6347 & 0.5904 & 0.7651 & 0.8186 & 0.8251 & \textbf{0.8425} & {0.8279} & \underline{0.8420} \\
			&LPIPS$\downarrow$ & 0.3947 & 0.3670 & 0.3386 & 0.2898 & 0.2811 & \underline{0.2723} & 0.2739 & \textbf{0.2291}\\
			&DISTS$\downarrow$ & 0.2201 & 0.1385 & 0.1776 &0.1468 & 0.1442 & \underline{0.1372} & 0.1406 & \textbf{0.1015} \\
			&NIQE$\downarrow$ & 7.536 & \underline{5.237} & 7.229 &5.528 & 5.340 & 5.242 & 5.392 & \textbf{5.212} \\
			&CLIP-IQA$\uparrow$  & 0.6817 & \textbf{0.7644} & 0.7365 & 0.7380 & 0.7410 & 0.7434 & 0.7411 & \underline{0.7451}\\
			\bottomrule[1pt]
		\end{tabular}
		\label{tab:reds_BI}
		\vspace{-0.2cm}
	\end{table*}
}

\subsection{Temporal Feature Alignment Module}
Frame-wisely conducting ISR models for video super-resolution could amplify the differences of blurry patterns \cite{shi2022rethinking} across frames, leading to content inconsistency such as the object shape deformation. 
The issue originated from solely relying on spatial level super-resolution and lacking temporal coherence modeling across frames.
To facilitate visual content alignment in video super-resolution, a temporal feature alignment (TFA) module is devised after each SFA module in UNet and VAE decoder, for the temporal feature interaction and calibration.

Figure \ref{fig:blocks}(d) depicts the learning procedure of TFA module.
Given the input HR intermediate feature maps $\tilde{F} = \{\tilde{f}^i\}_{i=1}^L$ from the SFA module in UNet decoder, we first partition the feature map $\tilde{f}^i$ of each frame into $N$ non-overlapping windows with the spatial resolution of $h\times{w}$. $N=\frac{HW}{hw}$ is the total window number.
Then, we link all features within a local window across $L$ frames to form a HR feature tubelet $\tilde{F}_{tub}\in\mathbb{R}^{L\times{h}\times{w}\times{C}}$.
We reshape the dimension of each HR feature tubelet into ${hwL\times{C}}$ and execute the standard self-attention on it: 
\begin{equation}
	\begin{aligned}
		&Q, K, V=\mathrm{Conv3D}(\tilde{F}_{tub}), \\
		&\hat{F}_{tub}= \mathrm{Attention}(Q,K,V),
	\end{aligned}
\end{equation} 
where $Q,K,V\in \mathbb{R}^{hwL\times{C}}$ are the \textit{query}, \textit{key} and \textit{value} matrices, respectively. 
Each of them is predicted by a 3D convolution layer. 
The self-attention conducted on the HR feature tubelet enables the feature interaction across different frames, mitigating the temporal feature misalignment in local regions. 
To further conduct the temporal feature calibration, we leverage the counterpart of HR feature tubelet, i.e., the feature tubelet ${G}_{tub}$ of the LR latent feature maps, as a reference for feature adjustment.
We perform the cross-attention between $\hat{F}_{tub}$ and ${G}_{tub}$ to obtain the output feature tubelet $\bar{F}_{tub}$: 
\begin{equation}
	\begin{aligned}
		&Q^\prime = \mathrm{Conv3D}(\hat{F}_{tub}),\quad K^\prime,V^\prime = \mathrm{Conv3D}({G}_{tub}), \\
		&\bar{F}_{tub}= \mathrm{Attention}(Q^\prime, K^\prime, V^\prime), 
	\end{aligned}
\end{equation}  
where the query $Q^\prime$ is learnt on the HR feature tubelet $\hat{F}_{tub}$ and the key/value $K^\prime$/$V^\prime$ is estimated on the LR counterpart via 3D convolution layers, respectively.
We collect all the output feature tubelets from the TFA module and reshape them into the original size as $\bar{F} \in \mathbb{R}^{L \times H \times W \times C}$.
The output feature $\bar{F}$ is then fed into the next block of the decoder in UNet or VAE for video latent denoising or reconstruction.

In this way, the coupled SFA and TFA modules in UNet and VAE decoder not only emphasize the pixel-wise feature adaptation for visual appearance preservation but also strengthen the temporally feature interaction and calibration for temporal coherence modeling.

\subsection{Video Refiner}  
Recent advance \cite{choi2022perception} reveals that images synthesized by diffusion model conditioning on visual contents might loss some original color information in local regions.
To address this problem, StableSR \cite{wang2023stablesr} performs a non-parametric post-processor to refine the generation with reference to original input for achieving color preservation.
Instead, we propose a trainable video refiner to emphasize the adjustment of decoded HR video from VAE decoder, by leveraging the information from up-sampled LR video.

Figure \ref{fig:blocks}(b) details the structure of our video refiner. 
We first concatenate the decoded video $X_d$ and the up-sampled LR video $X_u$ along channel dimension, and then feed it into a residual block.
The refined HR video $X_H$ is generated by fusing $X_u$, $X_d$ and the output feature mapping of residual block as: 
\begin{equation}
	X_H = w X_u + (1-w) X_d + ResBlock([X_u, X_d]),
\end{equation}
where $w$ is a trade-off parameter.
The devised video refiner balances the original visual contents of the up-sampled LR video and the synthesized contents of decoded HR video via feature fusion learning.   
Accordingly, our design is more powerful in terms of color preservation, and achieves a good trade-off between the synthesized quality and fidelity.

\subsection{Training Strategy} 
We construct our SATeCo for video super-resolution based on the Stable Diffusion \cite{rombach2022high} model. 
There are four training stages to optimize the whole architecture.
In the first stage, we train the video upscaler using the Charbonnier loss \cite{charbonnier1994loss} to optimize the video reconstruction of HR videos.  
After that, we follow the standard setting in \cite{rombach2022high} to train UNet for the optimization of the inserted SFA and TFA modules. 
We fix all parameters of UNet except for the two kinds of modules during training. 
For the optimization of SFA and TFA modules in VAE decoder, we take the video latent codes of the HR videos as the input, and optimize the similarity between the decoded videos and ground-truth HR videos.
Finally, we freeze all the parameters in video upscaler, UNet and VAE, and train the video refiner using the pairs of decoded and ground-truth HR videos.

\begin{figure*}[t]
	\centering
	\includegraphics[width=0.91\linewidth]{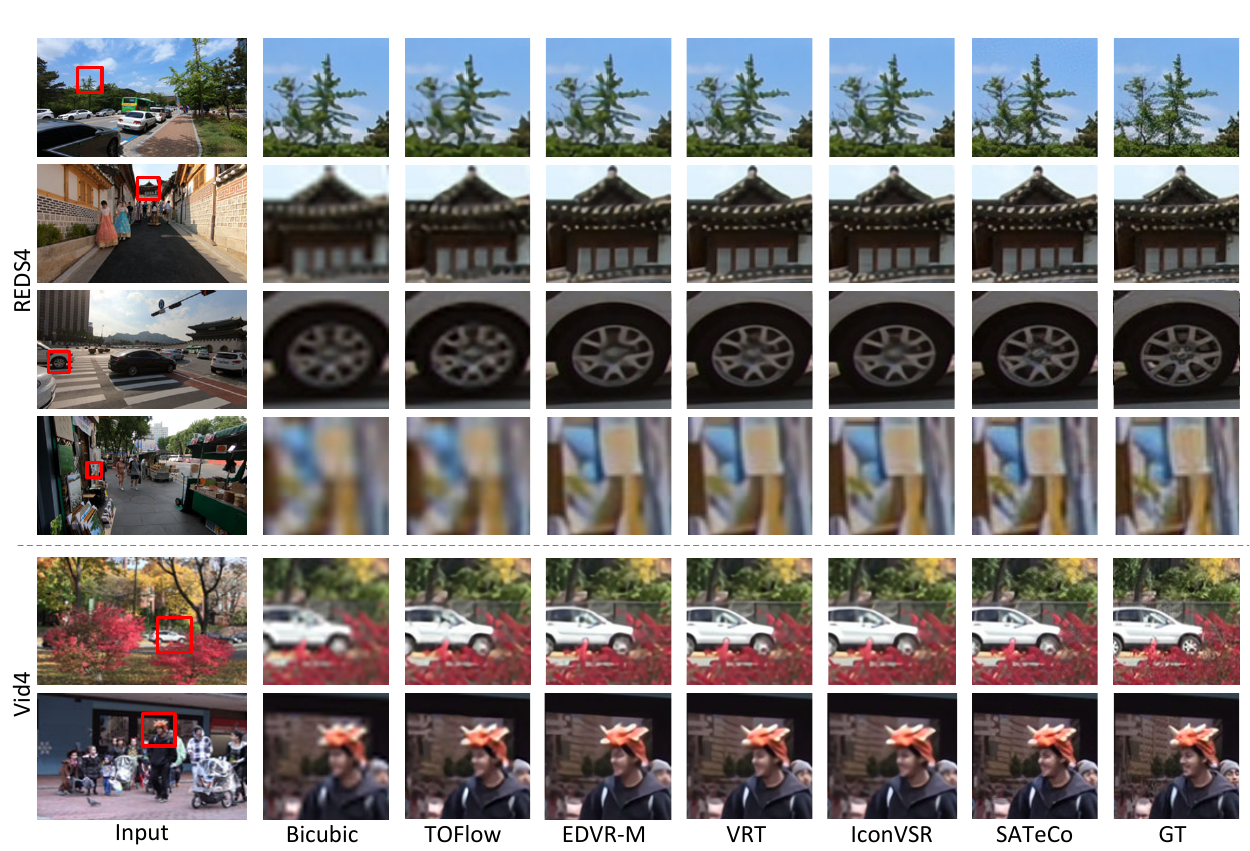}
	\vspace{-0.05in}
	\caption{Six visual examples of video super-resolution results by different approaches on the REDS4 and Vid4 datasets. The region in the red box is presented in the zoom-in view for comparison.}
	\label{fig:visualize}
	\vspace{-0.2in}
\end{figure*}

\section{Experiments}

\subsection{Experimental Settings}

\textbf{Datasets.}
We empirically evaluate the effectiveness of our SATeCo on two widely-used datasets: REDS \cite{nah2019ntire} and Vid4 \cite{liu2013bayesian}.
The \textbf{REDS} dataset consists of $240$, $30$ and $30$ video clips for training, validation and testing. 
Each video clip contains $100$ frames with the resolution of $1,280 \times 720$.
We employ the standard protocols in \cite{chan2021basicvsr,edvr2019,chan2022basicvsr++} and select four video clips from the validation set as the testing data, namely \textbf{REDS4}.
The \textbf{Vid4} dataset also includes four video clips, and there are about $40$ frames in each clip with the resolution of $720\times 480$.
Following the standard settings \cite{chan2022basicvsr++,liang2022vrt}, we employ all the videos in Vid4 for evaluation and choose the video data in the training set of Vimeo-90K \cite{xue2019flow} for model optimization.
There are $64,612$ training clips and each clip has 7 frames with the resolution of $448 \times 256$.

\textbf{Implementation Details.} 
We implement our SATeCo on the PyTorch platform by using Diffusers \cite{Diffusers} library. 
The noise scheduler is set as linear scheduler ($\beta_1=0.00085$, $\beta_T=0.0120$, and $T=1,000$).
The trade-off parameters $w$ in video refiner is determined as $0.5$ by cross validation. 
We empirically set the window size in TFA as $h=8$, $w=8$. 
The frame number $L$ of input clip is $6$. 
The model is trained with AdamW optimizer and the learning rate is $5.0\times 10^{-5}$.

\textbf{Evaluation Metrics.} 
We evaluate the VSR models via two kinds of metrics, i.e., pixel-based and perception-based metrics.
The pixel-based metrics include PSNR and SSIM which calculate the similarity of every pixel between the generated and ground-truth HR videos. 
There are also some perception-based evaluation metrics for super-resolution.
These metrics mainly measure video quality from the viewpoint of human perceptual preference, and we adopt LPIPS \cite{zhang2018unreasonable}, DISTS \cite{ding2020image}, NIQE \cite{mittal2012making} and CLIP-IQA \cite{wang2023exploring} in this paper.
Specifically, LPIPS utilizes VGG \cite{simon2015very} model to extract frame features and measures the feature similarity between the synthesized and ground-truth videos. 
DISTS also computes the feature similarity between video pairs via a variant of VGG model, but the emphasis is on image texture.
For NIQE and CLIP-IQA, the scores are directly predicted by the learnt models without using the ground-truth HR videos.
NIQE measures the similarity of feature distribution between synthesized frames and a realistic image set \cite{mittal2012making}, while CLIP-IQA computes cosine similarity between generated frames and text prompts (e.g., ``High Resolution'') via CLIP model \cite{radford2021learning}.
In addition, we conduct a user study to verify human preference on different models.

\begin{figure*}[t]
	\centering
	\includegraphics[width=0.93\linewidth]{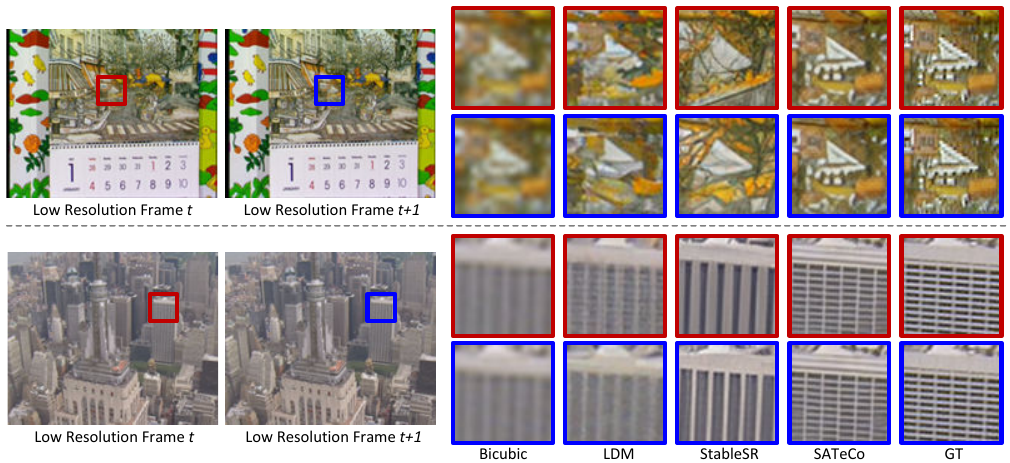}
	\caption{Video super-resolution results of two videos in the Vid4 dataset. The region in the same local position across two adjacent frames (i.e., regions highlighted by red and blue boxes) is scaled up to show more details.}
	\label{fig:consistency}
	\vspace{-0.2in}
\end{figure*}

\subsection{Comparisons with State-of-the-Art Methods}
We compare our SATeCo with several state-of-the-art techniques, including Bicubic Interpolation, StableSR~\cite{wang2023stablesr}, TOFlow~\cite{xue2019flow}, EDVR-M~\cite{edvr2019}, BasicVSR~\cite{chan2021basicvsr}, VRT~\cite{liang2022vrt} and IconVSR~\cite{chan2021basicvsr}, on REDS4 and Vid4 datasets.

\textbf{Quantitative Evaluation.}
Table \ref{tab:reds_BI} summarizes the performances of different VSR approaches in terms of the six metrics over the two datasets.
Overall, SATeCo achieves the best performances across all perception-based metrics (i.e., LPIPS, DISTS, NIQE and CLIP-IQA) on REDS4.
These metrics emphasize the quality judgment from human perceptual aspect and the results demonstrate the advantage of exploiting abundant knowledge prior in the pre-trained diffusion models to generate high-quality HR videos with better visual perception. 
In terms of pixel-based metrics, recent advances \cite{wang2023stablesr,yang2023pasd} manifest that the stochasticity in diffusion models could hurt the preservation of visual appearance in HR videos, resulting in inferior performances to traditional regression models.
Our SATeCo, by capitalizing on pixel-wise guidance from LR videos to modulate HR frame feature synthesis, alleviates the downsides and obtains the PSNR of $31.62$dB.
Notably, such performance is very comparable to that of IconVSR \cite{chan2021basicvsr}, which is the SOTA baseline of regression VSR models. 
The performance trends on Vid4 are similar with those on REDS4.
In particular, SATeCo attains the DISTS of $0.1015$, which relatively reduces that of the best competitor VRT \cite{liang2022vrt} by $26.0\%$.
The results indicate that SATeCo benefits from learning pixel-wise spatial adaption in diffusion to preserve frame-wise image texture for achieving better video fidelity.

\begin{figure}[t]
	\centering
	\includegraphics[width=0.85\linewidth]{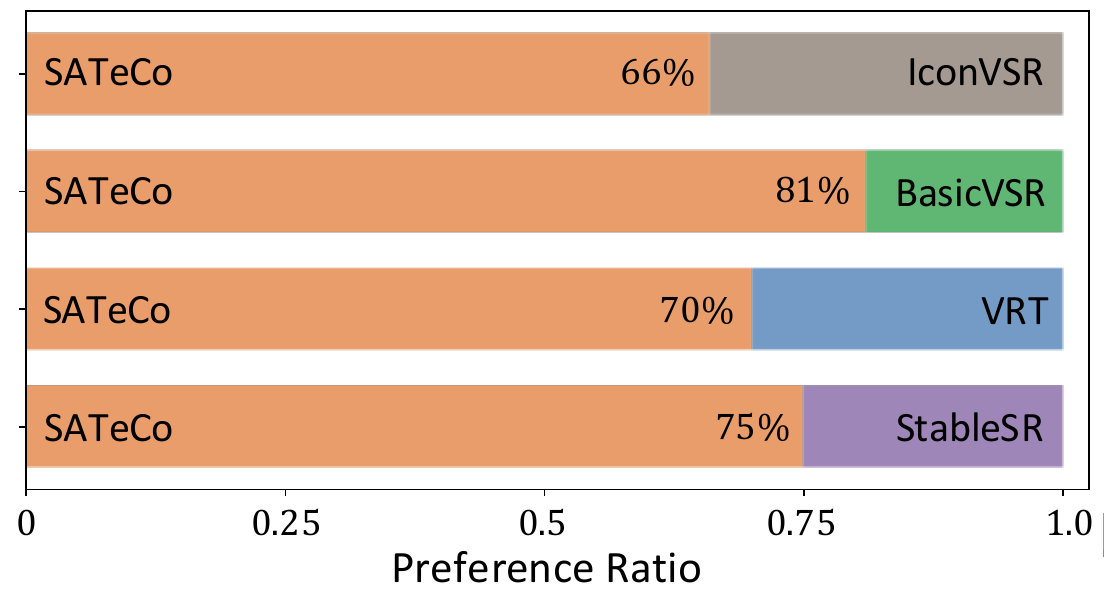}
	\vspace{-0.1in}
	\caption{Human evaluation of user preference ratios between SATeCo and other baselines on REDS4 and Vid4.}
	\label{fig:user}
	\vspace{-0.2in}
\end{figure}

\textbf{Qualitative Evaluation.}
Figure \ref{fig:visualize} visualizes the video super-resolution with six examples from REDS4 and Vid4.
Compared to other baselines, SATeCo can successfully restore more local details (e.g., the sharp edges in the eave and spoke of the 2nd and 3rd case) in frames with high fidelity.
Even with large blurriness (e.g., the 4th case), SATeCo still exhibits the strong restoration ability for video super-resolution, which again confirms the effectiveness of leveraging rich knowledge prior of diffusion models and learning spatial adaptation.
To further validate the temporal coherence learnt by SATeCo, we visualize two adjacent frames of two synthesized HR videos by using different diffusion-based super-resolution approaches in Figure \ref{fig:consistency}.
As observed in the figure, LDM and StableSR synthesize different visual contents across the two frames, e.g., the small windows in the building.
In contrast, our SATeCo predicts the HR videos with higher frame consistency and preserves the visual fidelity.
That basically validates the merit of performing tubelet-based self-attention within HR videos and cross-attention between HR videos and LR counterparts to achieve better temporal feature interaction and calibration.

\textbf{Human Evaluation.} 
Next, we further conduct human study to verify the HR video generation quality by using different VSR approaches with respect to the user preference.
We invite 100 evaluators on the Amazon MTurk platform, and ask each evaluator to choose the better one from two synthetic HR videos generated by two different methods given the same LR video.
Figure \ref{fig:user} depicts the user preference ratios on all eight videos in the REDS4 and Vid4 datasets.
SATeCo clearly wins the traditional regression models of IconVSR, BasicVSR and VRT, and the diffusion model of StableSR.
The results indicate SATeCo nicely magnifies LR videos with better visual quality and temporal coherence through the spatial feature adaptation and temporal feature alignment design in video diffusion procedure.

\setlength{\tabcolsep}{0.25mm}{
	\begin{table}\small
		\caption{Performance comparisons on REDS4 among variants with different integration of SFA and TFA modules.} 
		\vspace{-0.2cm}
		\centering
		\renewcommand\arraystretch{1.1}
		\begin{tabular}{l|c|c|c|c|c|c|c|c|c}
			\toprule[1pt]
			\multirow{2}{*}{Model} & \multicolumn{2}{c|}{UNet} & \multicolumn{2}{c|}{VAE} & \multirow{2}{*}{PSNR$\uparrow$} & \multirow{2}{*}{SSIM$\uparrow$} & \multirow{2}{*}{LPIPS$\downarrow$} & \multirow{2}{*}{DISTS$\downarrow$} & \multirow{2}{*}{NIQE$\downarrow$} \\
			\cline{2-5}
			& SFA & TFA & SFA & TFA &&&&&\\
			\midrule
			A& &&&  & 28.56 & 0.7925 & 0.2159 & 0.0758 & 4.404 \\ 
			B& $\checkmark$ &&&  & 28.93 & 0.8087 & 0.2042 & 0.0693 & 4.349 \\
			C& $\checkmark$ & $\checkmark$&&  & 29.45 & 0.8398 & 0.1892 & 0.0620 & 4.324 \\
			D& $\checkmark$ & $\checkmark$ &$\checkmark$& $\checkmark$ & \textbf{31.62} & \textbf{0.8932} & \textbf{0.1735} & \textbf{0.0607} & \textbf{4.104} \\
			\bottomrule[1pt]
		\end{tabular}
		\label{tab:stfc}
		\vspace{-0.25cm}
	\end{table}
}

\subsection{Model Analysis} 

\begin{figure}[t]
	\centering
	\includegraphics[width=1.0\linewidth]{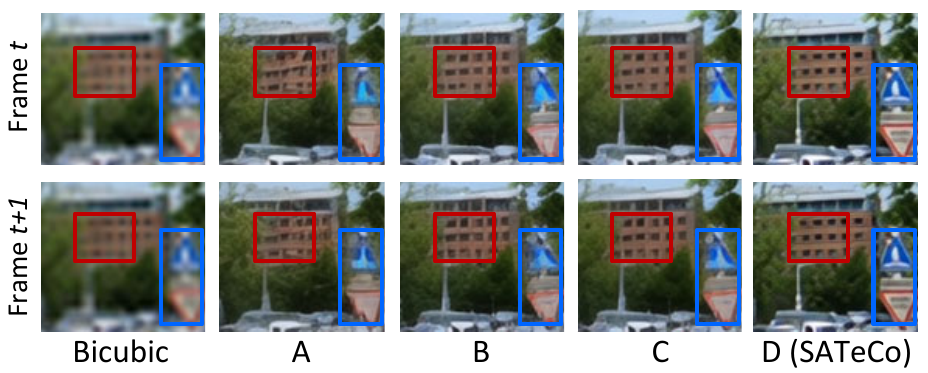}
	\vspace{-0.2in}
	\caption{Zoom-in view of two adjacent frames in one video super-resolution result synthesized by variants of SATeCo.}
	\label{fig:ablation}
	\vspace{-0.15in}
\end{figure}

\textbf{Analysis on SFA and TFA modules.}
We first investigate how the SFA and TFA modules influence the overall performances of video super-resolution.
Table \ref{tab:stfc} lists the performance comparisons among variants with different integration ways of SFA and TFA modules. 
We start from the basic diffusion model \textbf{A}, which leverages the zero-initialized convolution \cite{zhang2023adding} in UNet/VAE to learn the spatial guidance from LR videos for super-resolution. 
The model \textbf{B} and \textbf{C} gradually upgrade the basic model A through plugging SFA and TFA modules into the UNet, which improves the PSNR from $28.56$dB to $29.45$dB.  
Compared to zero-initialized convolution which simply conducts weighted summation of LR frame features and HR ones to guide spatial-level diffusion learning, the combination of SFA and TFA not only enhances the spatial adaptation via feature modulation but also strengthens temporal feature alignment by the tubelet-based attention. 
As such, the higher PSNR and SSIM which measure the spatial fidelity is attained by the model C. 
Finally, the model \textbf{D}, i.e., our SATeCo, by further exploiting SFA and TFA in VAE to regulate pixel-space video reconstruction, shows the best performances in PSNR and SSIM. 
In view of perception-based evaluation metrics, SATeCo also constantly obtains improvements over other variants, indicating the potential benefit from spatial-temporal guidance learning to enhance visual perception in HR videos.
Furthermore, Figure \ref{fig:ablation} showcases video super-resolution in a local region of one example in two adjacent frames.
SATeCo reconstructs the HR videos with high-quality visual appearance and promising temporal consistency among adjacent frames, proving the impact of exploring feature adaptation and alignment in diffusion for super-resolution.

\setlength{\tabcolsep}{0.7mm}{
	\begin{table}\small
		\caption{Ablation studies on the design of video upscaler and video refiner in SATeCo. The performances are reported on REDS4.} 
		\vspace{-0.05in}
		\centering
		\begin{tabular}{l|c|c|c|c|c|c}
			\toprule[1pt]
			\multicolumn{2}{c|}{Model} & PSNR$\uparrow$ & SSIM$\uparrow$ & LPIPS$\downarrow$ & DISTS$\downarrow$ & NIQE$\downarrow$ \\
			\midrule
			\multirow{2}{*}{Upscaler}& PixelShuffle & 29.77 & 0.8426 & 0.1979 & 0.0720 &4.298 \\
			& Ours & \textbf{31.62} & \textbf{0.8932} & \textbf{0.1735} & \textbf{0.0607} & \textbf{4.104} \\
			\midrule
			\multirow{3}{*}{Refiner}& $w=0$ & 30.36 & 0.8572 & \textbf{0.1581} & \textbf{0.0339} & \textbf{3.457} \\
			& $w=0.5$ & \textbf{31.62} & \textbf{0.8932} & 0.1735 & 0.0607 & 4.104 \\
			& $w=1.0$ & 28.99 & 0.8001 & 0.1815 & 0.0652 & 4.488 \\
			\bottomrule[1pt]
		\end{tabular}
		\label{tab:upscaler}
		\vspace{-0.2in}
	\end{table}
}

\textbf{Analysis on Video Upscaler.}
Then, we study the effectiveness of the video upscaler in the SATeCo.
One alternative is to employ the pre-trained Pixel Shuffle layer \cite{shi2016real} as the video upscaler.
The upper part of Table \ref{tab:upscaler} details the performances of the two approaches on REDS4.
Our approach exhibits better performances against PixelShuffle across all evaluation metrics, especially in PSNR and SSIM.
Technically, PixelShuffle resamples videos via directly conducting a 2D convolution layer on the input frames.
Instead, ours delves into the formulation of frame-wise correlation through temporal mutual self-attention, which is more effective in pixel feature enhancement for video resampling.
As such, ours effectively preserves the visual contents in LR videos and facilitates the subsequent video diffusion.

\textbf{Analysis on Video Refiner.}
The video refiner in SATeCo aims for adjusting the decoded HR videos from diffusion model by referring up-sampled original LR videos to alleviate color degradation.
The trade-off parameter $w$ in video refiner balances the impact of the visual contents between the decoded videos and the LR videos.
To evaluate the influence of the parameter $w$, we list the VSR performances by varying $w$ in the lower part of Table \ref{tab:upscaler}.
When $w$ is $0$, the performances on perception-based metrics are the best, but there is a slight performance drop on PSNR and SSIM.
The performances indicate that the synthesized visual contents by diffusion models are more acceptable for human visual system.  
In contrast, employing a large value of $w$ (e.g., $1.0$) for video refinement considers the information of LR videos more and weakens the contribution of diffusion models, affecting the quality of visual content generation. 
Therefore, we empirically set the $w$ as $0.5$ to seek a good trade-off between the synthesized contents and original visual appearance.

\section{Conclusions}
We have presented SATeCo that explores spatial adaptation and temporal coherence in diffusion models for video super-resolution.
In particular, we study the problem of learning spatial-temporal guidance from low-resolution videos to calibrate high-resolution video diffusion procedure.
To materialize our idea, SATeCo freezes all the parameters in the pre-trained UNet/VAE, and plugs the spatial feature adaptation (SFA) and temporal feature alignment (TFA) modules in each decoder block to regulate latent-space video denoising and pixel-space video reconstruction.
Through learning affine parameters on the guidance of low-resolution videos, SFA modulates the high-resolution features of each pixel to achieve spatial adaptation.
TFA performs self-attention within a tubelet to enhance feature interaction and further conducts cross-attention between the tubelet and its low-resolution counterpart to guide temporal feature alignment learning.
Experiments conducted on two video datasets, i.e., REDS4 and Vid4, validate the effectiveness of the proposed SATeCo for video super-resolution in terms of both spatial fidelity and temporal consistency.

\textbf{Acknowledgments.} This work was supported in part by the National Natural Science Foundation of China under Contract 62021001, and in part by the Fundamental Research Funds for the Central Universities under contract WK3490000007. It was also supported by the GPU cluster built by MCC Lab of Information Science and Technology Institution of USTC.

{
    \small
    \bibliographystyle{ieeenat_fullname}
    \bibliography{main}
}

\end{document}